\documentclass[11pt,a4paper]{article}
\usepackage{authblk}
\usepackage[hyperref]{emnlp2018}
\usepackage{times}
\usepackage{latexsym}
\usepackage{pgfplots}
\usepackage{url}
\usepackage{graphicx} 
\usepackage{amsmath,amssymb}
\usepackage{ifthen}
\usepackage{color}
\usepackage{multirow}
\usepackage{algorithm}
\usepackage[noend]{algpseudocode}
\algrenewcommand\algorithmicrequire{\textbf{Input:}}
\algrenewcommand\algorithmicensure{\textbf{Output:}}
\usepackage{enumitem}

\usepackage{url}

\aclfinalcopy 

\newcommand\sC{\ensuremath{\mathcal{C}}}

\newcommand\sK{\ensuremath{\mathcal{K}}}

\newcommand\R{\ensuremath{\mathbb{R}}} 

\ifthenelse{\isundefined{\definition}}{\newtheorem{definition}{Definition}}{}
\ifthenelse{\isundefined{\assumption}}{\newtheorem{assumption}{Assumption}}{}
\ifthenelse{\isundefined{\proposition}}{\newtheorem{proposition}{Proposition}}{}
\ifthenelse{\isundefined{\theorem}}{\newtheorem{theorem}{Theorem}}{}
\ifthenelse{\isundefined{\lemma}}{\newtheorem{lemma}{Lemma}}{}
\ifthenelse{\isundefined{\corollary}}{\newtheorem{corollary}{Corollary}}{}
\ifthenelse{\isundefined{\alg}}{\newtheorem{alg}{Algorithm}}{}
\ifthenelse{\isundefined{\example}}{\newtheorem{example}{Example}}{}

\newcommand\nl[1]{{\it``#1''}} 

\newcommand\zl[1]{\text{\footnotesize{\tt #1}}} 

\newcommand\reverse[1]{\mathbf R[#1]}

\title{Decoupling Structure and Lexicon for Zero-Shot Semantic Parsing}

  \author[1]{\textbf{Jonathan Herzig}}
\author[1,2]{\textbf{Jonathan Berant}}
\affil[1]{Tel-Aviv University}
\affil[2]{Allen Institute for Artificial Intelligence}
\affil[ ]{\tt {jonathan.herzig@cs.tau.ac.il, joberant@cs.tau.ac.il}}

\date{}

\begin{document}
\maketitle
\begin{abstract}
Building a semantic parser quickly in a new domain is a fundamental challenge
for conversational interfaces, as current semantic parsers require expensive
supervision and lack the ability to generalize to new domains. 
In this paper, we introduce a zero-shot approach to semantic parsing that can parse utterances in unseen domains while only being trained on examples in other source domains. 
First, we map an utterance to an abstract, domain-independent, logical form that 
represents the structure of the logical form, but contains slots instead of KB
constants. Then, we replace slots with KB constants via lexical alignment scores and global inference. 
Our model reaches an average accuracy of $53.4\%$ on $7$ domains in the \textsc{Overnight} dataset, substantially better than other zero-shot baselines, and performs as good as a parser trained on over $30\%$ of the target domain examples.
 
\end{abstract}

\section{Introduction}
\label{introduction}

Semantic parsing, the task of mapping natural language utterances into
executable logical forms, is a key paradigm in developing conversational interfaces 
\cite{zelle96geoquery,zettlemoyer05ccg,kwiatkowski11lex,berant2015agenda}.
The recent success of conversational interfaces such as Amazon Alexa, Google
Assistant, Apple Siri, and Microsoft Cortana has led to soaring interest in
developing methodologies for training semantic parsers quickly in any new
domain and from little data.

Prior work focused on alleviating data collection by training from weak
supervision
\cite{clarke10world,liang11dcs,kwiatkowski2013scaling,artzi2013weakly}, or developing protocols for fast data collection through paraphrasing
\cite{berant2014paraphrasing,wang2015overnight} or a human-in-the-loop
\cite{iyer2017learning}. However, all these approaches rely on supervised training data in the target domain and ignore data collected previously for other domains.

\begin{figure}[t]
  \includegraphics[width=1.0\columnwidth]{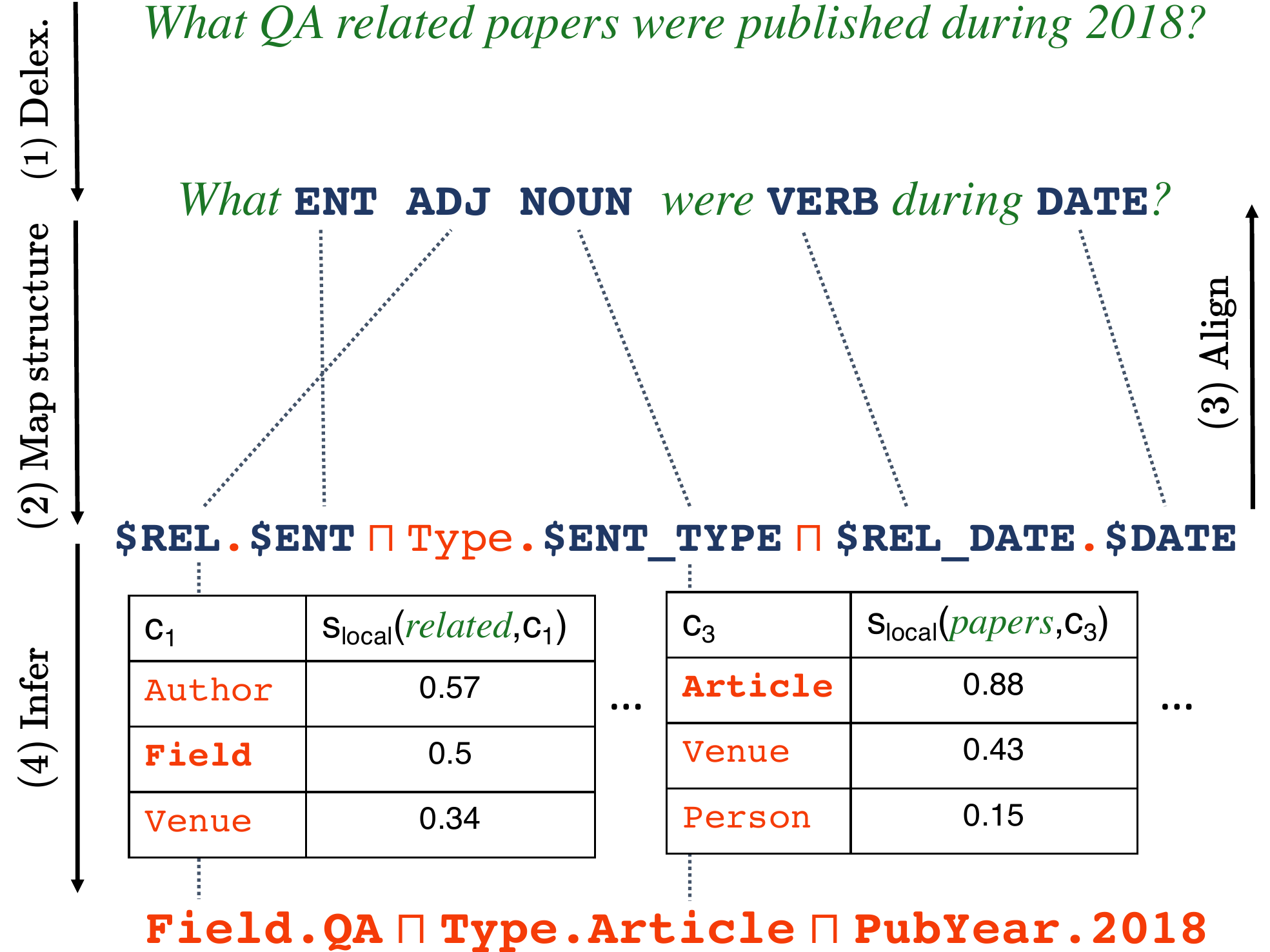}
  \caption{A test utterance is delexicalized (1) and mapped to its abstract
  logical form (2). Slots (``\$'' variables) are then aligned to the
  abstract utterance (3), and are filled with the top assignment in
  terms of local and global scores (4). Logical forms throughout this paper are
  in $\lambda$-DCS \cite{liang2013lambdadcs}.
  }~\label{fig:model}
\end{figure}

In this paper, we propose an alternative, zero-shot approach to semantic parsing, where no
labeled or unlabeled examples are provided in the target domain, but annotated examples 
from other domains are available. This is
a challenging setup as in semantic parsing
each dataset is associated with its own knowledge-base (KB) and thus all target domain KB constants (relations and entities) are unobserved at training time.
Moreover, this is a natural use-case as more and more conversational
interfaces are developed in multiple domains.

Our approach is motivated by recent work 
\cite{herzig2017multi,su2017cross,fan2017transfer,richardson2018polyglot} that showed that while the
lexicon and KB constants in different domains vary, the structure of language
composition repeats across domains. Therefore, 
we propose that by abstracting away the domain-specific lexical items
of an utterance, we
can learn to map the structure of an abstract utterance to an abstract
logical form that does not include any domain-specific KB constants, using data
from other domains only. 

Figure~\ref{fig:model} illustrates this approach. A test
utterance in the target domain is delexicalized and mapped to an abstract,
domain-independent representation, where some content words are replaced by 
abstract tokens (step 1).
Then, a structure-mapping model maps this representation
into an abstract logical form that contains slots instead of KB constants
(step 2).
A major
technical challenge at this point is to replace slots in the abstract logical
form with KB constants from the target domain. 
We show that it is possible to learn a domain-independent
lexical alignment model 
that aligns each slot to a word in the original utterance (step 3). This alignment,
combined with a global inference procedure (step 4) allows one to find the best assignment
of KB constants and produce a final logical form.
Importantly, both of our models are trained
from data in \emph{other} domains only.

We show that our zero-shot framework parses 7 different unseen domains from the
\textsc{Overnight} dataset with an average denotation accuracy of 53.4\%. This
result dramatically outperforms several natural baselines, and achieves the same
result as training a parser on over 30\% of the fully supervised target domain examples. To our knowledge, this work is the first to train a zero-shot semantic parser that can handle unseen domains. All our code is available at \url{https://github.com/jonathanherzig/zero-shot-semantic-parsing}.

\section{Background}
\label{sec:base_model}

\paragraph{Neural Semantic Parsing}

Sequence-to-sequence models~\cite{sutskever2014sequence} were recently proposed
for semantic
parsing~\cite{jia2016recombination,dong2016logical}. In
this setting, a sequence of input language tokens $x_1,\dots, x_m$ is
mapped to a sequence of output logical tokens $z_1,\dots,z_n$. We briefly review
the model by \newcite{jia2016recombination}, which we use as part of
our framework, and also as a baseline.

The \textbf{encoder} is a BiLSTM~\cite{hochreiter1997lstm} that converts $x_1,
\dots, x_m$ into a sequence of context sensitive states. 
The attention-based \textbf{decoder}~\cite{bahdanau2015neural,luong2015translation} is an LSTM language model additionally conditioned on the encoder states. Formally, the decoder is defined by:
\begin{align*}
  p(z_j &= w \mid x, z_{1:j-1}) \propto \exp(U[s_j, c_j]) , \\
  s_{j+1} &= LSTM([\phi^{(out)}(z_j),c_j],s_j) , \label{eq:state}
\end{align*}
where $s_j$ are decoder states, $U$ and the embedding function $\phi^{(out)}$
are the decoder parameters, and the context vector, $c_j$, is the result of global attention~\cite{luong2015translation}. 
We also employ attention-based copying \cite{jia2016recombination}, but omit details for brevity.

\paragraph{Semantic Parsing over Multiple KBs}
Recently, \newcite{herzig2017multi}, \newcite{su2017cross} and
\newcite{fan2017transfer} proposed to exploit structural regularities in
language across different domains. 
These works pooled together examples from multiple datasets in
different domains, each corresponding to a separate KB, and trained a single sequence-to-sequence model over
all examples, sharing parameters across domains. 
They showed that this substantially improves parsing accuracy.
While these works implicitly capture linguistic regularities across domains, they rely on annotated data in the target
domain. We, conversely, explicitly decouple structure mapping from the assignment of KB constants, and thus can tackle the zero-shot setting where
no target domain examples are available. This is the focus of the next
section.

\section{Zero-Shot Semantic Parsing}
\label{sec:zero_shot}

\subsection{Overview}

Following the empirical success of sharing structural information between
different semantic parsing domains, we propose in this paper to take a more
radical approach and to explicitly decouple semantic parsing into a structure
mapping model and a lexicon mapping model. We now provide an overview of our
approach and explain how this decoupling facilitates zero-shot semantic parsing.

We assume access to $D$ different source domains, where for
every domain $d$ we receive a KB $\sK_d$, and a training set of pairs of utterances and logical forms $\{(x_i,
z_i)\}_{i=1}^{N_d}$.
We further 
assume a lexicon $L$ that maps each KB constant in $\sK_d$ to a short phrase that describes it (e.g., $L$(\zl{PubYear})$\rightarrow$\nl{publication year}), as in
\newcite{wang2015overnight}. 
Finally, we assume a pre-trained, static, embedding function $\phi(w) \in \R^f$
for every word $w$, used to measure cross-domain lexical semantic similarity. Our goal is to train a semantic parser 
that maps a new utterance $x$ to the correct
logical form $z$ from a new domain $d_\text{new}$ given $\sK_{d_\text{new}}$.

\begin{figure}[t]
  \includegraphics[width=0.9\columnwidth]{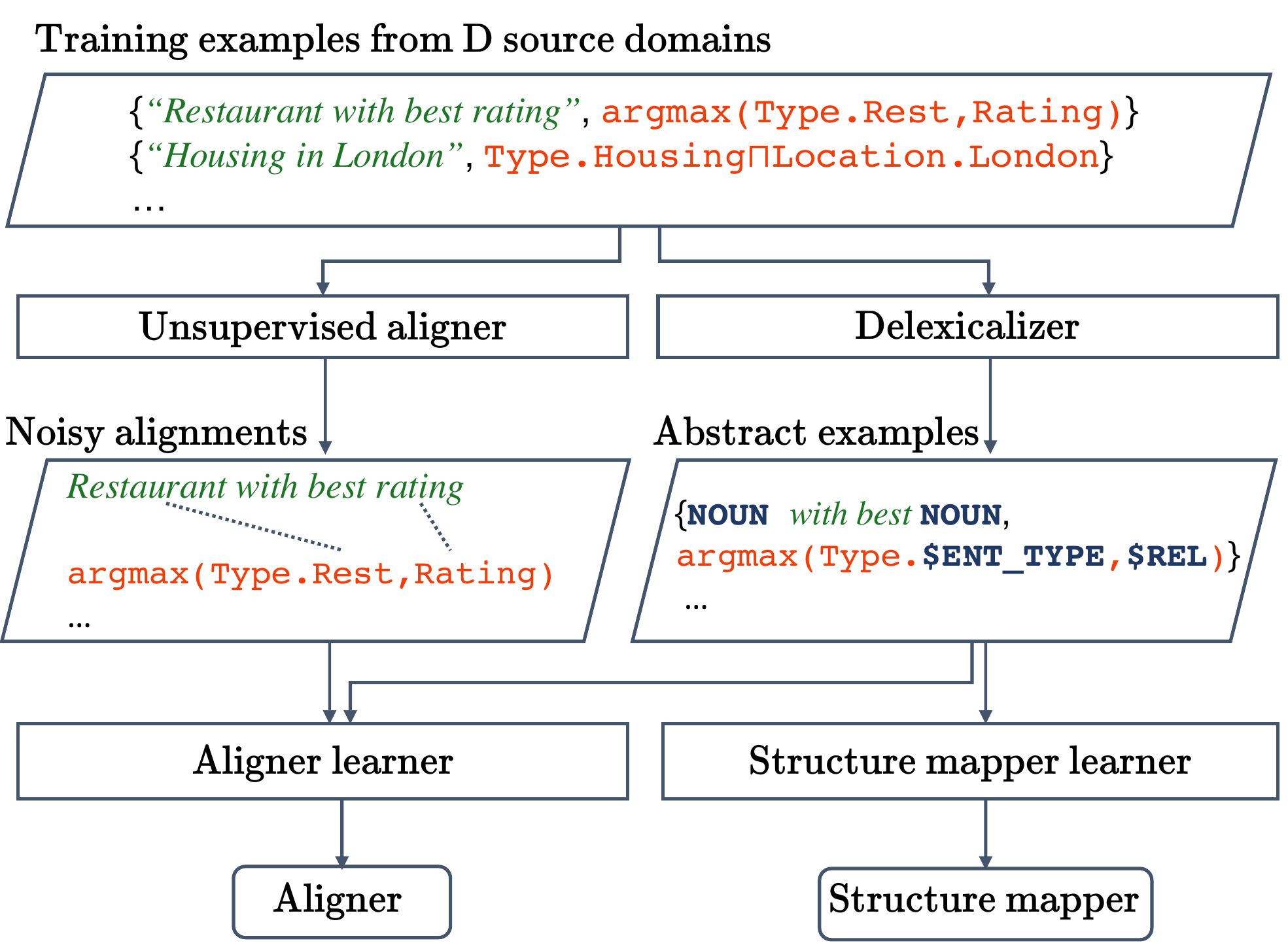}
  \caption{Training flow. Examples in source domains are delexicalized. Abstract examples are used to train both the structure mapper learner and aligner learner, where the aligner learner uses noisy alignments as labels.
  }~\label{fig:train}
\end{figure}

Figure~\ref{fig:train} describes the flow of our training procedure: we first employ
a simple rule-based method to transform training examples to an abstract
representation, where content words (in utterances) and KB constants (in logical forms) are delexicalized. We then
train the following two models that decouple structure from lexicon (a) The structure mapper that maps
abstract utterances to abstract logical forms. (b) The aligner that 
provides an alignment from abstract logical form tokens
to abstract utterance tokens.
Training the aligner is
challenging because no gold alignments between the abstract utterance
and abstract logical form are available. To overcome this challenge we propose a
distillation strategy: we obtain noisy supervision by training a
state-of-the-art unsupervised alignment model on the $D$ source
domains. Then, we train a second supervised alignment model that receives abstract
utterances, abstract logical forms, and target noisy alignments as input and learns to predict the
noisy alignments.

Once the two models are trained, we can tackle a new domain without training
examples (Figure~\ref{fig:model}). Given an utterance from the target domain,
we first abstract it using the delexicalizer, and then predict its abstract structure using the structure mapper. 
We treat delexicalized logical form tokens as slots to be filled with KB
constants. Candidate assignments are then scored locally according to the
semantic similarity of a KB constant (represented by its entry value in the lexicon $L$) to words the slot aligns to according to
the aligner. For this we use the pre-trained embedding function $\phi(\cdot)$ as the only cross-domain information.
Finally, we choose a final assignment of KB constants by exactly maximizing a global
scoring function, which takes into account both local alignment scores as well
as global constraints. 

We next describe in detail the four components of our framework: the delexicalizer,
structure mapper, aligner, and inference procedure.

\subsection{Delexicalizer}

\begin{figure}[t]

  {\footnotesize
	\setlength{\fboxsep}{0pt}
	\fbox{
		\parbox{\columnwidth}{
		
	\vspace{0.2cm}
		
    \textbf{Lexical representation}
		
    \nl{What meetings have no more than 3 attendees?}
    
    $\zl{Type.Meeting} \sqcap \reverse{\lambda x.\text{count}(\zl{Attendee}.x$)}.\zl{$\leq$.3}
    
	\nl{Which recipe needs no more than two ingredients?}   
	
	$\zl{Type.Recipe} \sqcap \reverse{\lambda x.\text{count}(\zl{IngredientOf}.x$)}.\zl{$\leq$.2}
    
    \vspace{0.2cm}
    \textbf{Abstract representation}
		
	\nl{What \texttt{NOUN} have no more than \texttt{NUM NOUN}?}
	
	     $\zl{Type.\$ENT\_TYPE} \sqcap \reverse{\lambda x.\text{count}(\zl{\$REL}.x$)}.\zl{$\leq$.\$NUM}
	
	\nl{Which \texttt{NOUN} \texttt{VERB} no more than \texttt{NUM NOUN}.}
	
	$\zl{Type.\$ENT\_TYPE} \sqcap \reverse{\lambda x.\text{count}(\zl{\$REL}.x$)}.\zl{$\leq$.\$NUM}

  	\vspace{0.1cm}
		
		}}
  }
  \caption{Examples in different domains (\textsc{Calendar} and \textsc{Recipes}) in their original and abstract representations. A similar structural regularity (a
comparative structure) maps to an identical abstract logical form.}
	\label{fig:shared}
\end{figure}

The goal of the delexicalizer is to strip utterances and logical forms from their domain-specific components and preserve domain-independent parts. 
We note that it is possible that some words contain both domain-specific and
domain-general aspects (\nl{cheapest}). However, we
conjecture that it is possible to decompose examples in a manner that
enables zero-shot semantic parsing.

The output of the delexicalizer is an abstract representation that should
manifest structural linguistic regularities across domains (Figure~\ref{fig:shared}). For example, 
a comparative structure will correspond to the same abstract logical form
in different domains.
In this representation, used as input to our models, content words and
KB constants are transformed to an abstract type. This rule-based preprocessing step is applied to all $D$
source domain training examples (utterances and logical forms), and to target domain
utterances at test time. We now describe the process of delexicalization.

\paragraph{Utterances}
Table~\ref{tab:delex} describes the full list of abstraction rules. 
We delexicalize several categories of content words and
keep function words, which describe the utterance structure, in their lexicalized
form.
Specifically, any verb\footnote{Numbers, dates and part-of-speech tags are extracted using Stanford
CoreNLP~\cite{manning2014stanford}.} whose lemma is not \nl{be} or \nl{do} is delexicalized.
All nouns are delexicalized, except for a small vocabulary of three words
(\nl{average}, \nl{total}, and \nl{number}), which denote a
domain-general operation. 
Adjectives tend to distribute more evenly between domain-specific words and domain-general words,
thus discriminating them is harder (e.g., \nl{outdoor}, \nl{wide} and
\nl{cooking} are domain-specific words while \nl{minimum}, \nl{same} and \nl{many} are
domain-general words). Thus, we take a statistical approach and only delexicalize
adjectives that are unique to the domain (i.e., did not appear in the training
set of any other source domain).
We also delexicalize dates and numbers, and identify entities in the utterance
by string matching against the entities in the KB. These are then delexicalized
to their corresponding abstract type (Table~\ref{tab:delex}).

\paragraph{Logical Forms}
We delexicalize all KB constants to their abstract type, which is given as part
of the KB schema (Table~\ref{tab:delex}).  

\begin{table}[t]
\centering
\resizebox{1.0\columnwidth}{!}{
\begin{tabular}{l|l|l|l}
Source       & Category           & Abstract Type & Examples                                    \\\hline\hline
Utterance    & Noun               & \zl{NOUN}          & \nl{cuisines}, \nl{housing}, \nl{time} \\
             & Verb               & \zl{VERB}          & \nl{published}, \nl{born}, \nl{posted}                     \\
             & Adjective          & \zl{ADJ}           & \nl{high}, \nl{cooking},  \nl{monthly}                     \\
             & Number             & \zl{NUM}           & \nl{4}, \nl{three}                                    \\
             & Date               & \zl{DATE}          & \nl{2018}, \nl{january 2nd}                           \\
             & Entity             & \zl{ENT}           & \nl{midtown}, \nl{alice}, \nl{dinner}                      \\\hline
Logical Form & Number             & \zl{\$NUM}         & \zl{1,2,3}                                       \\
             & Date               & \zl{\$DATE}        & \zl{1\_6\_2018}                                  \\
             & Entity             & \zl{\$ENT}         & \zl{MidtownWest, CentralOffice}                  \\
             & Entity type        & \zl{\$ENT\_TYPE}   & \zl{Person, Location, Recipe}                    \\
             & Numerical entity   & \zl{\$ENT\_NUM}    & \zl{Rent, Size, CookingTime}                     \\
             & Binary relation    & \zl{\$REL}         & \zl{Author, Attendee}                            \\
             & Unary relation     & \zl{\$REL\_UNARY}  & \zl{AllowsCats, WonAward}                        \\
             & Numerical relation & \zl{\$REL\_NUM}    & \zl{Height, Length, StarRating}                  \\
             & Date relation      & \zl{\$REL\_DATE}   & \zl{PostingDate, PubYear} 
\end{tabular}}
\caption{Categories of content words and KB constants, and their corresponding abstract type notation.}
\label{tab:delex}
\end{table}

\subsection{Structure Mapper}

As a first step towards predicting the lexical logical form, we map an abstract utterance, to an abstract logical form. The model is the neural semantic parser described in Section~\ref{sec:base_model}, only here the input and output are the abstract examples in all $D$ domains, which the delexicalizer outputs. The model utilizes a single encoder-decoder pair shared across all domains. 
As Figure~\ref{fig:shared} suggests, the model should learn, e.g., that a noun modified by a wh-question often maps to \zl{\$ENT\_TYPE},
and that \nl{no more than} maps to the $\leq$ operator.

\subsection{Aligner}

The output of the structure mapper is an abstract logical form that contains
slots instead of KB constants. To predict a complete logical form, we must assign a KB constant to each slot.

We observe that the description of a KB constant that appears in the logical
form (\zl{Article}) is often semantically similar to some word in the utterance (\nl{paper}). Thus, we 
can obtain signal for the identity of a KB constant by solving an alignment
problem: each slot can be aligned to words in the utterance that have similar meaning to that of the gold KB constant. 
Naturally, in some cases a KB constant is not semantically similar to any utterance word (e.g., the relation \zl{Field} in Figure \ref{fig:model}),
which we will mediate by using a global inference procedure
(Section~\ref{subsec:inference}).

\begin{figure}[t]
  \includegraphics[width=1.0\columnwidth]{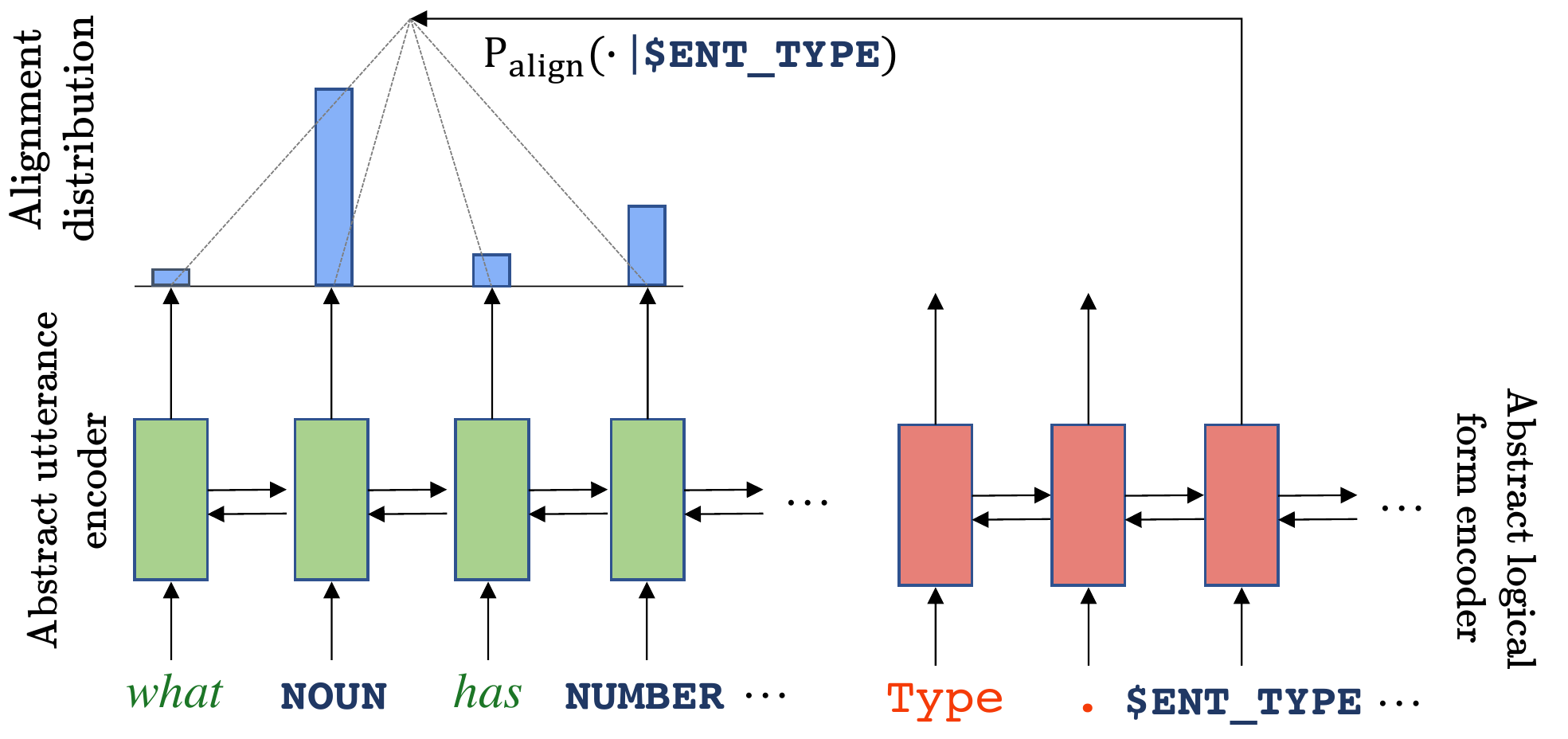}
\caption{The aligner model. Alignments are derived by comparing the slot hidden state against all utterance hidden states. 
  }~\label{fig:aligner}
\end{figure}

Thus, our goal is to learn a model that given an abstract utterance-logical form
pair $(x^{\text{abs}}, z^{\text{abs}})$ produces an alignment matrix $A$, where
$A_{ij}$ corresponds to the alignment probability $p(x_j^{\text{abs}} \mid
z_i^{\text{abs}})$. A central challenge is that no gold alignments are
provided in any domain. Therefore, we adopt a ``distillation approach", where we
train a supervised model over abstract examples to mimic the predictions of an
unsupervised model that has access to the full lexicalized examples.

Specifically, we use a standard unsupervised word aligner \cite{dyer2013simple}, which
takes all lexicalized examples $\{(x_i, z_i)\}_{i=1}^{N_d}$ in all $D$ domains and
produces an Alignment matrix $A^*$ for every example, where $A^*_{ij}=1$ iff token
$i$ in the logical form is aligned to token $j$ in the utterance. 
Then, we treat $A^*$ as gold alignments and generate examples $(x^{\text{abs}}, z^{\text{abs}}, A^*)$ to train
the aligner.
Learning alignments over abstract representations is possible, as a slot in a specific context tends to align to specific types of abstract words (e.g., Figure~\ref{fig:shared} suggests that a relation that is aggregated, often aligns to the \texttt{NOUN} that appears after the \texttt{NUM} 
in the abstract utterance). 

We now present our alignment model, depicted in Figure~\ref{fig:aligner}.
The model uses two different BiLSTMs to encode $x^{\text{abs}}$ and
$z^{\text{abs}}$ to their context sensitive states $b_1, \dots,b_m$ and
$s_1, \dots,s_n$ respectively. We model the alignment probability
$p_{\text{align}}(x^{\text{abs}}_j\mid z^{\text{abs}}_i)$ with a bi-linear form
similar to attention~\cite{luong2015translation}:
\begin{align*}
  e_{ij} &= s^T_iWb_j, \\
  p_{\text{align}}(x^{\text{abs}}_j\mid z^{\text{abs}}_i) &= \frac{\exp(e_{ij})}{\sum_{j'=1}^m \exp(e_{ij'})} ,
\end{align*}
where the parameters $W$ are learned during training. We train the model to
minimize the negative log-likelihood of gold alignments while considering only
alignments of slots (since we only align slots at test time). The cross-entropy loss for a
training example $(x^{\text{abs}},z^{\text{abs}}, A^*)$ is then given by:
\begin{equation*}
  -\sum^n_{i:i \in S_z} \sum^m_{j=1}A^*_{ij}\log p_{\text{align}}(x^{\text{abs}}_j\mid z^{\text{abs}}_i),
\end{equation*}
where $S_z$ are the slot indices in $z^{\text{abs}}$.

Our model can be viewed as an attention model, dedicated to aligning logical
form tokens to utterance tokens.
Using a separate alignment model rather than the attention weights of the
structure mapper has two advantages: First, alignments are generated given the
entire generated sequence $z^{\text{abs}}$ rather than just a prefix. Second, our
model focuses its capacity on the alignment task without worrying
about generation of $z^{\text{abs}}$. In Section~\ref{sec:experiments}, we will
demonstrate that training a dedicated aligner substantially improves performance.

\subsection{Inference} \label{subsec:inference}

The aligner provides a distribution over utterance tokens for every slot
in the abstract logical form. To compute the final logical form, we must replace
each slot with a KB constant. Formally, 
let $(z^\text{abs}_{j_1}, \dots, z^\text{abs}_{j_l})$ be the sequence of slots
in $z^\text{abs}$ and denote them for simplicity as $y=(y_1, \dots, y_l)$.
Our goal is to predict a sequence of KB constants $c=(c_1, \dots, c_l)$, where
each $c_i$ is chosen from a candidate set $\sC(y_i)$ that is determined by the abstract
token $y_i$ according to Table~\ref{tab:delex} 
(e.g.,
if $y_i$ is \zl{\$REL}, then $\sC(y_i)$ is the set of binary relations).

Our scoring function depends on alignments computed by the
aligner. However, because slots are
independent in the aligner, we introduce a few global constraints that capture
the dependence between different slots.
Formally, we wish to find $c^*$ that maximizes the following scoring function,
which depends on the utterance $x$, the slot sequence $y$, the abstract logical
form $z^\text{abs}$, the alignment matrix $A$ and the embedding function $\phi$:
\begin{align*}
\arg\max_c  \sum_{i=1}^l \big( s_\text{local}(c_i, y_i, x, A, \phi)\big) + s_\text{global}(c, z^\text{abs}).
\end{align*}
We now describe our scoring functions in detail.

\paragraph{Local Score}
Because inference is applied only at test time, we have access 
to the
lexicalized utterance and not only the abstract one.
Thus, the aligner outputs a distribution over words for a slot $y$ (e.g., in
Figure~\ref{fig:model}, \zl{\$REL\_DATE} aligns with high probability to
\texttt{VERB}, which corresponds to the word \nl{published}). 
Each word, in turn, has different semantic similarity to each KB constant in $\sC(y)$.
Intuitively, we would like to assign a KB constant that has high similarity with words the slot is aligned to. Thus, 
we define $s_\text{local}$ of a KB constant $c_i$ for every slot $y_i$ to be its
expected semantic similarity under the alignment distribution:
\begin{align*}
  &s_{\text{local}}(c_i, y_i, x, A, \phi) = E_{x \sim
  p_{\text{align}}(x^{\text{abs}} \mid y_i)} [\emph{sim}_\phi(x, c_i)] \\
  &= \sum_{j=1}^m p_{\text{align}}(x^{\text{abs}}_j \mid y_i) \cdot \emph{sim}_\phi(x_j,c_i).
\end{align*}

We define $\emph{sim}_\phi(x_j, c_i)$ to be the cosine similarity between 
the embedding $\phi(x_j)$ and the embedding $\phi(c_i)$ (scaled to the range [0,1]), where
$\phi(c_i)$ is defined to be the average embedding of all words in $L(c_i)$,
that is, $\emph{sim}(x_j,c_i)=\frac{1 + \cos(\phi(x_j), \phi(c_i))}{2}$.

\paragraph{Global Score}
Utilizing only a local scoring function raises several concerns. First, slots are
treated independently and dependencies between slots are ignored, which might
result in a final logical form that is globally inconsistent. For example, we
could generate the logical form \zl{Birthplace.ComputerScience}, which is
semantically dubious.
Second, some KB constants do not align to any word in the utterance and 
appear in the logical form only implicitly.
For example, the logical form in Figure \ref{fig:model} contains the \zl{Field} relation, however \nl{field} is implicit in the utterance. 
Therefore, we define $\emph{exe}_\sK(z)$ to be true iff $z$ executes against
$\sK$ without errors, and define a global score that prevents assignments $c$ that result
in a logical form $z$ such that $\emph{exe}_\sK(z)$ is false.

Moreover, we can use similar constraints to prevent logical forms that are
highly unlikely according to our prior knowledge. Specifically, we define
$\emph{once}(z)$ to be true iff each date, named entity, and number in the
logical form $z$ appear exactly once. We then define a global score that
prevents logical forms in which $\emph{once}(z)$ is false.
Empirically, we find such assignments to be
mostly wrong
(e.g., \zl{Type.Article $\sqcap$ (Field.QA $\sqcup$ Field.QA)}).

Formally, our scoring function is defined as:
\begin{align*}
  s_{\text{global}}(c, z^{abs}) = \left\{ \begin{array}{rl}
    0 &\mbox{ \small $\emph{exe}_\sK(z^{abs}|_c)$, $\emph{once}(z^{\text{abs}}|_c)$} \\
-\infty &\mbox{ otherwise,}
\end{array} \right.
\end{align*}
where $z^{abs}|_c$ is the result of assigning the KB constants $c$ to the slots
in $z^{abs}$.

\paragraph{Inference Algorithm.} 
While each local scoring function can be efficiently maximized independently,
the global constraints that depend on the entire assignment $c$ make inference
more complicated. However, because the global scoring function introduces hard 
constraints, an exact and efficient inference algorithm is still possible.
Our inference algorithm generates solutions one-by-one sorted by the
local scoring function only. Then, it checks for each one whether it satisfies
the global constraints defined by $s_\text{global}$, and stops once a satisfying
solution is found, which is guaranteed to maximize our scoring function. While
in the worst case, this procedure is exponential in the size of $c$, in
practice solutions are found after only a few steps. 
We also always halt after $T$ steps if a solution has not been found. 

Algorithm~\ref{alg:infer} describes the details of our inference procedure.
We define $\emph{cands}$ to be a data structure that contains $l$ lists of
candidate KB constants (a list for each slot), sorted according to the local
scoring function $s_\text{local}$ in descending order. Additionally,
$\emph{getAssign}(\emph{cands}, a)$ is a function that accesses $\emph{cands}$,
and retrieves the assignment with indices $a$. For example,
$\emph{getAssign}(\emph{cands},  \{0\}^l)$ retrieves the top scoring local assignment. Last, we define $a_{\text{inc}(i)}$ to be the indices $a$, where $a_i$ is incremented by $1$.

\begin{algorithm}[t]
  {\footnotesize
\caption{Exact inference algorithm}\label{alg:infer}
\begin{algorithmic}[1]
  \Require $\emph{cands}$, $T$
\Ensure $c^*$ - the top scoring assignment
  \State $\emph{horizon} \gets \emptyset$\Comment{Max heap}
\State $a_{\text{init}} \gets \{0\}^l$
  \State $\emph{push}(\emph{horizon},  a_{\text{init}}$)
\For{$t\gets 1$ \textbf{to} $T$}
  \State $a\gets \emph{pop}(\emph{horizon})$
  \State $c\gets \emph{getAssign}(\emph{cands},a)$
  \If{$s_\text{global}(c, z^{\text{abs}})=0$}
\State \textbf{return} $c$
\EndIf
\For{$i\gets 1$ \textbf{to} $l$}
  \State $\emph{push}(\emph{horizon}, a_{\text{inc}(i)})$
\EndFor
\EndFor
\State \textbf{return} $NULL$
\end{algorithmic}
  }
\end{algorithm}

The algorithm proceeds as follows. First we initialize a maximum heap
\emph{horizon} into which we will dynamically push candidate assignments.
Then, we iteratively pop the best current assignment from the heap, and check if
it satisfies the global constraints. If it does, we return this assignment and stop. Otherwise, we generate the next possible candidates, one
from each list (there is no need to add more than one because candidates are
sorted). If no satisfying assignment is found after $T$ steps, we return $NULL$.
It is easy to show that when the algorithm returns an assignment it is
guaranteed to be the one that maximizes our global scoring function.

\section{Experiments}
\label{sec:experiments}


\begin{table*}[t]
  {\scriptsize
\centering
\resizebox{1.0\textwidth}{!}{
\begin{tabular}{l|ccccccc|c}
\hline\hline
Model        & Blocks & Calendar & Housing & Publications & Recipes & Restaurants & Social & Avg.  \\\hline
\textsc{InLex}         & 59.9 & 73.8 & 72.0 & 79.5 & 79.2 & 76.2 & 83.4 & 74.8 \\
\textsc{InAbstract}       & 39.6 & 57.7 & 51.9 & 59.6 & 66.2 & 68.4 & 66.1 & 58.5 \\
\textsc{CrossLex}      & 0.0  & 0.0  & 0.5  & 0.0  & 1.4  & 1.2  & 1.0  & 0.6  \\
\textsc{CrossLexRep} & 6.5  & 1.8  & 2.6  & 1.9  & 16.7 & 6.6  & 1.7  & 5.4  \\
\textsc{ZeroShot}    & 28.3 & 53.6 & 52.4 & 55.3 & 60.2 & 61.7 & 62.4 & 53.4
\end{tabular}}
\caption{Test accuracy for all models on all domains. 
  }
\label{tab:res}
}
\end{table*}

\subsection{Experimental Setup}

\paragraph{Data}
We evaluated our method on the \textsc{Overnight} semantic parsing dataset, which contains $13,682$ examples of language utterances paired with logical
forms across eight domains, which were chosen to explore diverse types of language phenomena. 
As described, our approach depends on having linguistic regularities repeat across
domains. However two domains contain logical forms that are based on
neo-davidsonian semantics for treating events with multiple arguments. Since such
logical forms are completely absent in six domains, it is not possible for our
method to generalize to those in our zero-shot approach. Therefore, we do not
evaluate on the 
\textsc{Basketball} domain, in which 98\% of the examples contain such logical
forms, and omit all examples (68\%) that contain such logical forms in the
\textsc{Social} domain.
We evaluated on the same train/test split as \newcite{wang2015overnight}, using
the same accuracy metric, i.e., the proportion of questions for which the denotations of the predicted and gold logical forms are equal. 
We additionally used the lexicon $L$ they provided with descriptions for KB constants.

\paragraph{Evaluated Models}
We evaluated different models (Table~\ref{tab:models}) according to the
following two attributes. Firstly, whether the model is trained on target domain
data (\textit{in-domain}) or on source domains data only (\textit{cross-domain}). Secondly,
we trained the neural semantic parser described in Section~\ref{sec:base_model} over the lexical data representation (\textit{lexical}), or in comparison trained our model over the abstract representation (\textit{abstract}).

\begin{table}[t]
\centering
\resizebox{0.8\columnwidth}{!}{
\begin{tabular}{l|l|l}
         & \textbf{In-domain} & \textbf{Cross-domain} \\\hline
\textbf{Lexical}  & \textsc{InLex}         & \textsc{CrossLex}            \\\hline
\textbf{Abstract} & \textsc{InAbstract}         & \textsc{ZeroShot}    
\end{tabular}}
\caption{Evaluated models.}
\label{tab:models}
\end{table}

As \textsc{CrossLex} can not generate KB constants unseen during training, we additionally implemented \textsc{CrossLexRep}. In this model, we added an additional step that modifies the output of \textsc{CrossLex}: we replaced a generated KB constant with its most similar KB constant from the target KB that also shares its abstract type. 

\paragraph{Implementation Details}
In all experiments, for our embedding function $\phi(\cdot)$, we used pre-trained GloVe~\cite{pennington2014glove} vectors with dimension $300$. 
In a single experiment we considered one domain as the target domain, while
other domains were the source domains (and repeated for all domains).
For \textsc{InLex}, \textsc{CrossLex} and \textsc{CrossLexRep} we used exactly
the same experimental setup as~\newcite{jia2016recombination}. For our zero-shot
model, we used 20\% of the training data as a development set for tuning
hyper-parameters. We first tuned parameters for the structure mapper, and used
the best setting for tuning the aligner.

We provide the list of hyper-parameters and their values for our zero-shot framework. \textbf{Structure mapper:} number of epochs ($22$, using early stopping), hidden unit dimension (300), word vector dimension ($100$), learning rate ($0.1$ with SGD optimizer), $L2$ regularization ($0.001$). At test time, we used beam search with beam size 5, and then picked the highest-scoring logical form that we could infer an assignment for.
\textbf{Aligner:} number of epochs (30, using early stopping), hidden unit dimension (250), word vector dimension (100), learning rate ($0.0002$ with Adam optimizer), dropout rate over hidden states ($0.4$). For both models, word vectors are updated during training.
\textbf{Inference:} we used $T=500$ steps, after which we halted.

\begin{table*}[t]
  {\scriptsize
\centering
\resizebox{1.0\textwidth}{!}{
\begin{tabular}{l|ccccccc|c}
\hline\hline
Model        & Blocks & Calendar & Housing & Publications & Recipes & Restaurants & Social & Avg.  \\\hline
\textsc{ZeroShot}            & 29.2 & 60.4 & 57.3 & 47.7 & 59.5 & 63.0 & 64.0 & 54.5 \\\hline
\textsc{-GlobalHeur}          & 29.2 & 60.4 & 56.7 & 46.9 & 60.1 & 46.0 & 62.2 & 51.6 \\
\textsc{-Aligner}        & 22.6 & 57.5 & 52.7 & 46.9 & 58.4 & 43.0 & 60.4 & 48.8 \\
\textsc{-Inference}      & 28.2 & 51.5 & 44.7 & 43.0 & 42.2 & 44.5 & 45.8 & 42.8 \\
\textsc{-Aligner,Inference} & 20.4 & 38.8 & 32.7 & 37.5 & 37.0 & 15.5 & 36.9 & 31.2
\end{tabular}}
\caption{Development accuracy for all ablations. 
}
\label{tab:abl}
  }
\end{table*}

\subsection{Results}
We trained all models above and evaluated on the test set for all seven domains. Results show (Table~\ref{tab:res}) that \textsc{ZeroShot} substantially
outperforms other zero-shot baselines. \textsc{CrossLex} performs poorly, as it
can only generate KB constants seen during training. \textsc{CrossLexRep}
performs better, as it can generate KB constants from the target domain,
however, generating the correct constant usually fails. This highlights the challenge in the zero-shot semantic parsing setting.   

For baselines trained on target domain data, \textsc{InLex} (re-implementation
of~\cite{jia2016recombination}) achieved average accuracy of 74.8,
which is comparable to the 74.4 average accuracy they report on our seven
domains. 
Training on the target domain with our method \textsc{InAbstract} achieved
58.5\% average accuracy, 
which shows that while the abstract representation in our framework loses some
valuable information, it is still successful. Importantly, the 
performance of \textsc{ZeroShot} (53.4\%) is only slightly lower than \textsc{InAbstract},
showing that our model degrades gracefully and 
generalizes well across domains
compared to \textsc{CrossLex}.

\paragraph{Model Ablations}
We now measure the effect of different components of our framework on denotation accuracy. We examined
the effect of removing components completely, or replacing them with simpler
ones. Thus, the following ablated models can be viewed as additional baselines.
\begin{enumerate}[topsep=0pt,itemsep=0pt,parsep=0pt,wide=0pt, widest=99,leftmargin=\parindent, labelsep=*]
\itemsep0em 
\item \textsc{-Aligner}: Replacing the alignment distribution from the aligner with
  alignment distribution from the decoder of the structure mapper. 
\item \textsc{-Inference}: Discarding the global scoring function and maximizing
  each slot independently.
  \item \textsc{-Aligner,Inference}: Discarding both of our main technical
  contributions.
\item \textsc{-GlobalHeur}: Discarding the global inference heuristics (denoted as $once(z)$). 
\end{enumerate}
Table~\ref{tab:abl} shows that ablating each of the components hurts
performance.
Discarding our two main technical contributions results in 31.2\% accuracy
compared to 54.5\% in the full model. 
Performing inference with global constraints dramatically improve performance, showing that using the
alignment model alone results often in incoherent logical forms. 
Our dedicated aligner also improves performance compared to alignments learned by the decoder of the structure mapper. 
This is pronounced without global constraints (a drop from 42.8\% to
31.2\%), but is less severe when global inference is used (a drop from 54.5\% to
48.8\%).

\paragraph{Intrinsic Analysis}
While we evaluated performance above via denotation accuracy, we now evaluate
our framework's modules with different metrics (on the development set).
We evaluated the structure mapper by measuring the exact match of the top candidate in the beam to the gold abstract logical form ($49.1\%$). We further evaluated the aligner by measuring alignment accuracy for top candidate alignments, in comparison to the unsupervised aligner output ($72.9\%$). 

Finally, we measured inference performance in the following ways. The fraction of cases where inference succeeded within $T$ steps is $70\%$ (as some predicted abstract logical forms are not valid in terms of their syntax), and the average number of steps in case of success ($3.67$ steps). In addition, the fraction of correct global assignments given an abstract logical form that exactly matches the gold one is $77.0\%$. 
To conclude, results show that the structure mapping problem is harder than slot filling, for which we learned good alignments and performed fast and mostly accurate inference.

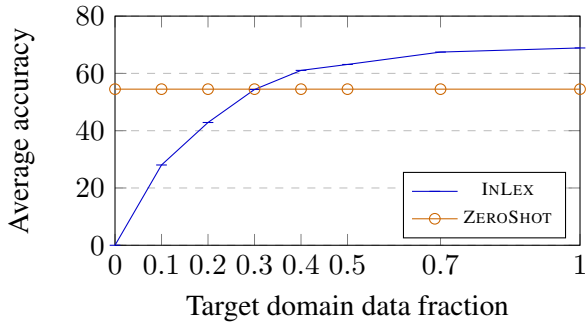
\begin{figure}[t]
\begin{tikzpicture}
\begin{axis}[
	width = 1.0\columnwidth,
	height = 0.6\columnwidth,
    xlabel={Target domain data fraction},
    ylabel={Average accuracy},
    xmin=0, xmax=1,
    ymin=0, ymax=80,
    xtick={0, 0.1, 0.2, 0.3, 0.4, 0.5, 0.7, 1.0},
    legend pos=south east,
    legend style={font=\scriptsize},
    ymajorgrids=true,
    grid style=dashed,
]
   
\addplot
	[
    color=blue!80!black,
    mark=-,
    ]
    coordinates {
    (0.0,0)(0.1,28.026)(0.2,42.859)(0.3,54.385)(0.4,61.048)(0.5,63.121)(0.7,67.446)(1,68.884)
    };
    \addlegendentry{\textsc{InLex}}
\addplot
	[
    color=orange!80!black,
    mark=o,
    ]
    coordinates {
    (0.0,54.5)(0.1,54.5)(0.2,54.5)(0.3,54.5)(0.4,54.5)(0.5,54.5)(0.7,54.5)(1,54.5)
    };
    \addlegendentry{\textsc{ZeroShot}}
\end{axis}
\end{tikzpicture}
\caption{Learning curve for \textsc{InLex}, compared to \textsc{ZeroShot} average performance.}
\label{fig:learn_curve}
\end{figure}

\paragraph{Valuation}
To estimate the value of our zero-shot framework in terms of target domain
examples, we plot a learning curve (Figure \ref{fig:learn_curve}) that shows
development set average accuracy for \textsc{InLex} (trained on target domain data). 
In comparison, \textsc{ZeroShot} utilizes no target domain data, thus it is
fixed. As Figure~\ref{fig:learn_curve} shows, our framework's value is equal to
30\% of the target domain training data. In our setting this equals to $400$ 
examples manually-annotated with full logical forms. Note that this value is
gained every time a semantic parser for a new domain is needed. Moreover,  
our parser can be used as an initial system, deployed to begin
training from user interaction directly.

\paragraph{Limitations}
We now outline some of the limitations of our approach for zero-shot semantic
parsing.
We hypothesized that language regularities repeat across domains, however as
mentioned above, neo-davidsonian semantics occurs mostly in one domain in the
\textsc{Overnight} dataset and thus we were not able to generalize to it. 
Our parser also obtained low accuracy in \textsc{Blocks}. 
This domain contains mostly spatial language, different from other domains in \textsc{Overnight}. 
Specifically, 
prepositions, which we did not lexicalize map to relations in the KB  (e.g., \nl{below} and \nl{above} map to the relations \zl{Below} and \zl{Above}). 
This shows the challenge involved in decomposing the structure from the lexicon
with rules.
In addition, since some spatial relations in this domain are semantically
similar (\zl{Length}, \zl{Width} and \zl{Height}), 
we found it hard to rank them correctly during inference.
This stresses that in our framework, we assume KB constants to be sufficiently
distinguishable in the pre-trained embedding space, which is not always the case.

\section{Related Work}
\label{sec:related_work}

While zero-shot executable semantic parsing is still under-explored, some works focused on the open-vocabulary setting which handles unseen relations by replacing a formal KB with a probabilistic database learned from a text corpus \cite{choi2015scalable,gardner2017open}.

Our abstract utterance representation is related to other attempts to generate intermediate representations that improve generalization such as dependency trees \cite{reddy2016transforming}, syntactic CCG parses \cite{krishnamurthy2015learning}, abstract templates \cite{abujabal2017automated,goldman2017weakly} or masked enitites \cite{dong2016logical}. Our abstract logical form representation is similar to that \newcite{dong2018coarse} used in to guide the decoding of the full logical form. The main difference with our work is that we focus on a comprehensive abstract representation tailored for zero-shot semantic parsing. 

It is worth mentioning other work that inspected various aspects of zero-shot
parsing. \newcite{bapna2017towards} focused on frame semantic parsing, and
assumed that relations appear across different domains to learn a better mapping
in the target domain. Also in frame semantic parsing, \newcite{ferreira2015zero}
utilized word embeddings to map words to unseen KB relations. Finally,
\newcite{lake2017still} inspected whether neural semantic parsers can handle
types of compositionality that were unseen during training. 
The main difference between their work and ours is that we focus on a scenario
where a compositional logical form is generated, 
but the target KB constants do not appear in any of the source domains.  

\section{Conclusion}
\label{conclusion}
In this paper we address the challenge of zero-shot semantic parsing. We
introduce a model that can parse utterances in unseen domains by decoupling structure mapping from lexicon mapping, and demonstrate its success on $7$ domains from the \textsc{Overnight} dataset.

In future work, we would like to automatically learn a delexicalizer from data, 
tackle zero-shot parsing when the structure distribution in the target domain is
very different from the source domains, and  apply our framework to datasets
where only denotations are provided.

\section*{Acknowledgments}
We thank Kyle Richardson, Vivek Srikumar  and the anonymous reviewers for their constructive feedback. This work was completed in partial fulfillment for the PhD degree of the first author. Herzig was supported by a Google PhD fellowship. This research was partially supported by The Israel Science Foundation grant 942/16 and The Blavatnik Computer Science Research Fund.

\bibliography{all}

\begin{thebibliography}{35}
\expandafter\ifx\csname natexlab\endcsname\relax\def\natexlab#1{#1}\fi

\bibitem[{Abujabal et~al.(2017)Abujabal, Yahya, Riedewald, and
  Weikum}]{abujabal2017automated}
Abdalghani Abujabal, Mohamed Yahya, Mirek Riedewald, and Gerhard Weikum. 2017.
\newblock Automated template generation for question answering over knowledge
  graphs.
\newblock In \emph{Proceedings of the 26th international conference on world
  wide web}, pages 1191--1200. International World Wide Web Conferences
  Steering Committee.

\bibitem[{Artzi and Zettlemoyer(2013)}]{artzi2013weakly}
Y.~Artzi and L.~Zettlemoyer. 2013.
\newblock Weakly supervised learning of semantic parsers for mapping
  instructions to actions.
\newblock \emph{Transactions of the Association for Computational Linguistics
  (TACL)}, 1:49--62.

\bibitem[{Bahdanau et~al.(2015)Bahdanau, Cho, and Bengio}]{bahdanau2015neural}
D.~Bahdanau, K.~Cho, and Y.~Bengio. 2015.
\newblock Neural machine translation by jointly learning to align and
  translate.
\newblock In \emph{International Conference on Learning Representations
  (ICLR)}.

\bibitem[{Bapna et~al.(2017)Bapna, Tur, Hakkani-Tur, and
  Heck}]{bapna2017towards}
Ankur Bapna, Gokhan Tur, Dilek Hakkani-Tur, and Larry Heck. 2017.
\newblock Towards zero shot frame semantic parsing for domain scaling.
\newblock In \emph{Interspeech 2017}.

\bibitem[{Berant and Liang(2014)}]{berant2014paraphrasing}
J.~Berant and P.~Liang. 2014.
\newblock Semantic parsing via paraphrasing.
\newblock In \emph{Association for Computational Linguistics (ACL)}.

\bibitem[{Berant and Liang(2015)}]{berant2015agenda}
J.~Berant and P.~Liang. 2015.
\newblock Imitation learning of agenda-based semantic parsers.
\newblock \emph{Transactions of the Association for Computational Linguistics
  (TACL)}, 3:545--558.

\bibitem[{Choi et~al.(2015)Choi, Kwiatkowski, and
  Zettlemoyer}]{choi2015scalable}
Eunsol Choi, Tom Kwiatkowski, and Luke Zettlemoyer. 2015.
\newblock Scalable semantic parsing with partial ontologies.
\newblock In \emph{Proceedings of the 53rd Annual Meeting of the Association
  for Computational Linguistics and the 7th International Joint Conference on
  Natural Language Processing (Volume 1: Long Papers)}, volume~1, pages
  1311--1320.

\bibitem[{Clarke et~al.(2010)Clarke, Goldwasser, Chang, and
  Roth}]{clarke10world}
J.~Clarke, D.~Goldwasser, M.~Chang, and D.~Roth. 2010.
\newblock Driving semantic parsing from the world's response.
\newblock In \emph{Computational Natural Language Learning (CoNLL)}, pages
  18--27.

\bibitem[{Dong and Lapata(2016)}]{dong2016logical}
L.~Dong and M.~Lapata. 2016.
\newblock Language to logical form with neural attention.
\newblock In \emph{Association for Computational Linguistics (ACL)}.

\bibitem[{Dong and Lapata(2018)}]{dong2018coarse}
Li~Dong and Mirella Lapata. 2018.
\newblock Coarse-to-fine decoding for neural semantic parsing.
\newblock In \emph{Proceedings of the 56th Annual Meeting of the Association
  for Computational Linguistics (Volume 1: Long Papers)}, pages 731--742.
  Association for Computational Linguistics.

\bibitem[{Dyer et~al.(2013)Dyer, Chahuneau, and Smith}]{dyer2013simple}
Chris Dyer, Victor Chahuneau, and Noah~A. Smith. 2013.
\newblock A simple, fast, and effective reparameterization of ibm model 2.
\newblock In \emph{Proceedings of the 2013 Conference of the North American
  Chapter of the Association for Computational Linguistics: Human Language
  Technologies}, pages 644--648, Atlanta, Georgia. Association for
  Computational Linguistics.

\bibitem[{Fan et~al.(2017)Fan, Monti, Mathias, and Dreyer}]{fan2017transfer}
Xing Fan, Emilio Monti, Lambert Mathias, and Markus Dreyer. 2017.
\newblock Transfer learning for neural semantic parsing.
\newblock In \emph{Proceedings of the 2nd Workshop on Representation Learning
  for NLP}, pages 48--56, Vancouver, Canada. Association for Computational
  Linguistics.

\bibitem[{Ferreira et~al.(2015)Ferreira, Jabaian, and
  Lef{\`e}vre}]{ferreira2015zero}
Emmanuel Ferreira, Bassam Jabaian, and Fabrice Lef{\`e}vre. 2015.
\newblock Zero-shot semantic parser for spoken language understanding.
\newblock In \emph{Sixteenth Annual Conference of the International Speech
  Communication Association}.

\bibitem[{Gardner and Krishnamurthy(2017)}]{gardner2017open}
Matt Gardner and Jayant Krishnamurthy. 2017.
\newblock Open-vocabulary semantic parsing with both distributional statistics
  and formal knowledge.
\newblock In \emph{AAAI}, pages 3195--3201.

\bibitem[{Goldman et~al.(2018)Goldman, Latcinnik, Nave, Globerson, and
  Berant}]{goldman2017weakly}
Omer Goldman, Veronica Latcinnik, Ehud Nave, Amir Globerson, and Jonathan
  Berant. 2018.
\newblock Weakly supervised semantic parsing with abstract examples.
\newblock In \emph{Proceedings of the 56th Annual Meeting of the Association
  for Computational Linguistics (Volume 1: Long Papers)}, pages 1809--1819.
  Association for Computational Linguistics.

\bibitem[{Herzig and Berant(2017)}]{herzig2017multi}
Jonathan Herzig and Jonathan Berant. 2017.
\newblock Neural semantic parsing over multiple knowledge-bases.
\newblock In \emph{Proceedings of the 55th Annual Meeting of the Association
  for Computational Linguistics (Volume 2: Short Papers)}, pages 623--628.
  Association for Computational Linguistics.

\bibitem[{Hochreiter and Schmidhuber(1997)}]{hochreiter1997lstm}
S.~Hochreiter and J.~Schmidhuber. 1997.
\newblock Long short-term memory.
\newblock \emph{Neural Computation}, 9(8):1735--1780.

\bibitem[{Iyer et~al.(2017)Iyer, Konstas, Cheung, Krishnamurthy, and
  Zettlemoyer}]{iyer2017learning}
Srinivasan Iyer, Ioannis Konstas, Alvin Cheung, Jayant Krishnamurthy, and Luke
  Zettlemoyer. 2017.
\newblock Learning a neural semantic parser from user feedback.
\newblock In \emph{Proceedings of the 55th Annual Meeting of the Association
  for Computational Linguistics (Volume 1: Long Papers)}, pages 963--973,
  Vancouver, Canada. Association for Computational Linguistics.

\bibitem[{Jia and Liang(2016)}]{jia2016recombination}
R.~Jia and P.~Liang. 2016.
\newblock Data recombination for neural semantic parsing.
\newblock In \emph{Association for Computational Linguistics (ACL)}.

\bibitem[{Krishnamurthy and Mitchell(2015)}]{krishnamurthy2015learning}
Jayant Krishnamurthy and Tom~M Mitchell. 2015.
\newblock Learning a compositional semantics for freebase with an open
  predicate vocabulary.
\newblock \emph{Transactions of the Association for Computational Linguistics},
  3:257--270.

\bibitem[{Kwiatkowski et~al.(2013)Kwiatkowski, Choi, Artzi, and
  Zettlemoyer}]{kwiatkowski2013scaling}
T.~Kwiatkowski, E.~Choi, Y.~Artzi, and L.~Zettlemoyer. 2013.
\newblock Scaling semantic parsers with on-the-fly ontology matching.
\newblock In \emph{Empirical Methods in Natural Language Processing (EMNLP)}.

\bibitem[{Kwiatkowski et~al.(2011)Kwiatkowski, Zettlemoyer, Goldwater, and
  Steedman}]{kwiatkowski11lex}
T.~Kwiatkowski, L.~Zettlemoyer, S.~Goldwater, and M.~Steedman. 2011.
\newblock Lexical generalization in {CCG} grammar induction for semantic
  parsing.
\newblock In \emph{Empirical Methods in Natural Language Processing (EMNLP)},
  pages 1512--1523.

\bibitem[{Lake and Baroni(2017)}]{lake2017still}
Brenden~M Lake and Marco Baroni. 2017.
\newblock Still not systematic after all these years: On the compositional
  skills of sequence-to-sequence recurrent networks.
\newblock \emph{arXiv preprint arXiv:1711.00350}.

\bibitem[{Liang(2013)}]{liang2013lambdadcs}
P.~Liang. 2013.
\newblock Lambda dependency-based compositional semantics.
\newblock \emph{arXiv}.

\bibitem[{Liang et~al.(2011)Liang, Jordan, and Klein}]{liang11dcs}
P.~Liang, M.~I. Jordan, and D.~Klein. 2011.
\newblock Learning dependency-based compositional semantics.
\newblock In \emph{Association for Computational Linguistics (ACL)}, pages
  590--599.

\bibitem[{Luong et~al.(2015)Luong, Pham, and Manning}]{luong2015translation}
M.~Luong, H.~Pham, and C.~D. Manning. 2015.
\newblock Effective approaches to attention-based neural machine translation.
\newblock In \emph{Empirical Methods in Natural Language Processing (EMNLP)},
  pages 1412--1421.

\bibitem[{Manning et~al.(2014)Manning, Surdeanu, Bauer, Finkel, Bethard, and
  McClosky}]{manning2014stanford}
C.~D. Manning, M.~Surdeanu, J.~Bauer, J.~Finkel, S.~J. Bethard, and
  D.~McClosky. 2014.
\newblock The stanford core{NLP} natural language processing toolkit.
\newblock In \emph{ACL system demonstrations}.

\bibitem[{Pennington et~al.(2014)Pennington, Socher, and
  Manning}]{pennington2014glove}
J.~Pennington, R.~Socher, and C.~D. Manning. 2014.
\newblock Glove: Global vectors for word representation.
\newblock In \emph{Empirical Methods in Natural Language Processing (EMNLP)}.

\bibitem[{Reddy et~al.(2016)Reddy, T{\"a}ckstr{\"o}m, Collins, Kwiatkowski,
  Das, Steedman, and Lapata}]{reddy2016transforming}
Siva Reddy, Oscar T{\"a}ckstr{\"o}m, Michael Collins, Tom Kwiatkowski, Dipanjan
  Das, Mark Steedman, and Mirella Lapata. 2016.
\newblock Transforming dependency structures to logical forms for semantic
  parsing.
\newblock \emph{Transactions of the Association for Computational Linguistics},
  4:127--140.

\bibitem[{Richardson et~al.(2018)Richardson, Berant, and
  Kuhn}]{richardson2018polyglot}
Kyle Richardson, Jonathan Berant, and Jonas Kuhn. 2018.
\newblock Polyglot semantic parsing in apis.
\newblock In \emph{Proceedings of the 2018 Conference of the North American
  Chapter of the Association for Computational Linguistics: Human Language
  Technologies, Volume 1 (Long Papers)}, pages 720--730. Association for
  Computational Linguistics.

\bibitem[{Su and Yan(2017)}]{su2017cross}
Yu~Su and Xifeng Yan. 2017.
\newblock Cross-domain semantic parsing via paraphrasing.
\newblock In \emph{Proceedings of the 2017 Conference on Empirical Methods in
  Natural Language Processing}, pages 1235--1246, Copenhagen, Denmark.
  Association for Computational Linguistics.

\bibitem[{Sutskever et~al.(2014)Sutskever, Vinyals, and
  Le}]{sutskever2014sequence}
I.~Sutskever, O.~Vinyals, and Q.~V. Le. 2014.
\newblock Sequence to sequence learning with neural networks.
\newblock In \emph{Advances in Neural Information Processing Systems (NIPS)},
  pages 3104--3112.

\bibitem[{Wang et~al.(2015)Wang, Berant, and Liang}]{wang2015overnight}
Y.~Wang, J.~Berant, and P.~Liang. 2015.
\newblock Building a semantic parser overnight.
\newblock In \emph{Association for Computational Linguistics (ACL)}.

\bibitem[{Zelle and Mooney(1996)}]{zelle96geoquery}
M.~Zelle and R.~J. Mooney. 1996.
\newblock Learning to parse database queries using inductive logic programming.
\newblock In \emph{Association for the Advancement of Artificial Intelligence
  (AAAI)}, pages 1050--1055.

\bibitem[{Zettlemoyer and Collins(2005)}]{zettlemoyer05ccg}
L.~S. Zettlemoyer and M.~Collins. 2005.
\newblock Learning to map sentences to logical form: Structured classification
  with probabilistic categorial grammars.
\newblock In \emph{Uncertainty in Artificial Intelligence (UAI)}, pages
  658--666.

\end{thebibliography}
\bibliographystyle{acl_natbib_nourl}


\end{document}